# *Automated Grading of Anatomical Objective Structured Practical Exams Using Decision Trees*

Jason Bernard[1], Ranil Sonnadara[1], Anthony N. Saraco[2], Josh P. Mitchell[2], Alex B. Bak[3], Ilana Bayer[2,4], Bruce C. Wainman[2,4].

[1]Department of Surgery, McMaster University, Hamilton, Ontario, Canada

[2]Education Program in Anatomy, Faculty of Health Sciences, McMaster University, Hamilton, Ontario, Canada

[3]Temerty Faculty of Medicine, University of Toronto, Toronto, Ontario, Canada

[4]Department of Pathology and Molecular Medicine, McMaster University, Hamilton, Ontario, Canada

Correspondence to: Dr. Bruce Wainman, Department of Pathology and Molecular Medicine, HSC 1R1, McMaster University, 1280 Main St. W. Hamilton ON, Canada, L8S 4K1.

Email: wainmanb@mcmaster.ca

ABSTRACT

An Objective Structured Practical Examination (OSPE) is an effective and robust, but resource-intensive, means of evaluating anatomical knowledge. Since most OSPEs employ short answer or fill-in-the-blank style questions, the format requires many people familiar with the content to mark the exams. However, the increasing prevalence of online delivery for anatomy and physiology courses could result in students losing the OSPE practice that they would receive in face-to-face learning sessions. The purpose of this study was to test the accuracy of Decision Trees (DTs) in marking OSPE questions as a potential first step to creating an intelligent, online OSPE tutoring system. The study used the results of the winter 2020 semester final OSPE from McMaster University's anatomy and physiology course in the Faculty of Health Sciences (HTHSCI 2FF3/2LL3/1D06) as the data set. Ninety percent of the data set was used in a 10-fold validation algorithm to train a DT for each of the 54 questions. Each DT was comprised of unique words that appeared in correct, student-written answers. The remaining 10% of the data set was marked by the generated DTs. When the answers marked by the DT were compared to the answers marked by staff and faculty, the DT achieved an average accuracy of 94.49% across all 54 questions. This suggests that machine learning algorithms such as DTs are a highly effective option for OSPE grading and are suitable for the development of an intelligent, online OSPE tutoring system.

INTRODUCTION

Medical education is shifting from a time-based model to a competency-based curriculum (Harris et al., 2020; Jalali et al., 2020). To grant the Doctor of Medicine (MD) degree, MD programs in Canada and the United States must meet the standards of accreditation set by the Liaison Committee on Medical Education (LCME) which requires defined professional competencies for students. In 2015, the Royal College of Physicians and Surgeons of Canada updated the educational framework used throughout Canadian medical schools, called CanMEDS, wherein medical students are now expected to develop competencies related to the multifaceted Role of Medical Expert such as Professional, Communicator, Collaborator, Leader, Health Advocate, and Scholar (Sherbino et al., 2011). Although the length of medical school has largely remained unchanged, undergraduate medical education is now further spread across studies in medical ethics, humanities, and professionalism as well as a movement towards increased and earlier clinical experiences (Doukas et al., 2015). Within this finite curricular space, dedicated time for anatomy education has suffered with a decline in students' contact time during undergraduate medical curriculum, in Canada and globally (Collins, 2008; Kramer et al., 2008; Leveritt et al., 2016; Mcbride and Drake, 2018; Rockarts et al., 2020). Anatomy education in most departments has been integrated with other preclinical courses with a progression towards clinically oriented anatomy using approaches such as problem-based learning with clinical cases, radiology, and targeted dissections (Stringer and Nicholson, 2008; McBride and Drake 2018; Rockarts et al. 2020). Despite being a frequent target for budgetary cuts, anatomy education facilitates the development of several core competencies necessary for becoming a medical expert (Hefler and Ramnanan, 2017).

This research aims to help with anatomical education by facilitating the grading of objective structured practical examinations (OPSEs) with an automated grading system using machine learning. Machine learning is a method to find complex patterns in data in a general way of finding a model analytically instead of building a model via explicit instructions or rules. For example, with medical imagery, machine learning has been used to recognize cancer cells by having the computer discover the commonalities of a cancerous region throughout set of images. For this research, it is hypothesized that there are patterns in the words used by the students to answer OSPE questions. To this end, decisions trees (DTs) were investigated as an algorithmic means of classifying an answer as correct or incorrect (Quinlan, 1986). While there is some benefit to automated grading by reducing workload, that was not the primary goal for this research (as it is useful for teaching assistants to grade exams to gain experience). This research was done as a first step towards constructing an intelligent tutoring and testing system for anatomical education and, possibly, as a precursor to other topics in medical education.

**Anatomy OSPEs in Competency-Based Medical Education**

Although teaching models have been diversified to adapt to this reduced contact time in anatomy (Jalali et al., 2020; Rockarts et al., 2020), assessments have largely been dependent on OSPEs to fulfill the needs of competency-based medical education (Yaqinuddin et al., 2013; Vishwakarma et al., 2016; Hefler and Ramnanan, 2017). The OSPE comprises multiple stations, with one student at each station at a time. Each station serves to test a specific structured objective by asking a single question or multiple questions on an associated anatomical specimen or model (Nayar et al., 1986). Questions are focused on a specific anatomical structure within the specimen, denoted by the placement of a colored pin, as well as relating that structure to function or imaging (Yaqinuddin et al., 2013). Students are given a fixed time limit to answer all questions at a given station. Once the time has elapsed, a bell rings, and each student rotates to their next station. The examination ends when all of the students have completed all stations.

**Resource-Intensiveness of OSPEs**

There are substantial logistic and economic burdens for conducting OSPEs in anatomy. Laboratory space must be cordoned amidst regular class schedules to prepare and administer each OSPE, and the necessary anatomical material also needs to be reserved (Daly, 2010). The constrained resources must then be delivered to a steadily increasing medical student class size (McBride and Drake, 2018). A greater percentage of the instructional staff's time must be dedicated, out of necessity, to manually grading the increasing number of exams. Therefore, the limited time of staff is being diverted away from teaching, research, or course development (Krippendorf et al., 2008).

There have been steps to mitigate the resource-intensiveness of grading an OSPE through digitization, while studying the impact on student performance. Efforts to transition an assessment that has traditionally been offered on pencil-and-paper to a computer-based platform suggested that student performance was not affected (Guimarães et al., 2018; Inuwa, 2012; Meyer et al., 2016), with one study (Guimarães et al., 2018) reporting familiarization with a computer-based system as an important first step. There have been other attempts to automate the workload, such as changing the nature of answers from free response to multiple choice (Shaibah and Vleuten, 2013) and to administer through portable tablets (Wilson et al., 2017). These studies, however, report limitations in cueing effects, resulting in significantly easier tests.

**Emergence of Artificial Intelligence in Assessments**

The study authors viewed machine learning as a potential tool to address issues concerning the resource-intensiveness of OSPE grading. In recent years, there has been a concerted effort to automate short answer grading in academia and industry. These automated systems have been met with criteria for success governed by their agreement with human markers and maximizing accuracy (Siddiqi et al., 2010). Natural Language Processing (NLP) is a popular field of research that many previous machine scoring algorithms draw upon. NLP is an intersection between artificial intelligence and linguistics that focuses on enabling computers to process, understand, and generate natural language. A machine scoring competition with Kaggle, a popular online data science community (Shermis, 2015), found that the most popular NLP technique used was Latent Semantic Analysis (LSA) (Dumais, 2004). LSA has shown to perform well with textual similarities independent of word meaning. However, achieving a perfectly accurate system is met with challenges, created by linguistic diversity through pronoun variations, negatives, and syntactic variety (Leacock and Chodorow, 2003). The short responses seen in OSPEs are often objectively evaluable as they benefit from questions which are highly specific and close-ended. Short answer questions differ from essays as more importance is placed on the amount of relevant content and less emphasis on writing ability, spelling, and/or mechanical errors (Shermis, 2015). OSPE responses are uniquely suited for automated grading using machine learning, otherwise called "machine scoring", because the answers are highly specific and usually short. In some cases, an OSPE question is even similar to a fill-in-the-blank question with single terms. The origins of anatomical terminology are often rooted in lemmas and supported by orienting words (e.g., anterior) which is conducive to NLP and textual similarity approaches. The hypothesis was that the machine scoring of OSPE exams would be a relatively simple task if an adequate algorithm could be employed.

BACKGROUND

**Course Information**

McMaster University's Health Sciences Human Anatomy and Physiology (HTHSCI 2F03/2FF3/2L03/2LL3/1D06/1H06) is a multidisciplinary undergraduate anatomy and physiology course. The course has approximately 900 students enrolled per year from the Bachelor of Health Sciences, Integrated Biomedical Engineering and Health Sciences, Engineering, Midwifery, and Nursing programs. All disciplines enrolled in the course take the same end of semester OSPE, except for the nursing students. The fall semester focuses on communication and locomotion, and thus coveres cells and tissues, muscle physiology, the nervous system, special senses, the musculoskeletal system, and the endocrine system. The winter semester focuses on homeostasis and focuses on the cardiovascular, immune, respiratory, gastrointestinal, and urogenital systems. The fall 2019 semester OSPE was performed in-person. However, the winter 2020 semester OSPE was performed online due to COVID-19. Therefore, data from the winter semester online OSPE was used for this study.

Ethics

The protocol for this study was deemed exempt from Hamilton Integrated Research Ethics Board (HiREB) ethics review, as it constituted anonymous data gathered for the purposes of course improvement. The data to be obtained did not require opinions but was simply available as program metrics. On that basis the project was granted a waiver of ethics review, as per the Tri-Council Policy Statement (TCPS 2; Articles 2.1, 2.2, and 2.5), (TCPS, 2018).

**Traditional OSPE Grading Approach in the Education Program in Anatomy**

McMaster University's Health Sciences anatomy andphysiology course (HTHSCI 2F03/2FF3/2L03/2LL3/1D06) OSPEs are typically marked in a conference style, colloquially referred to as the "marking bee". Faculty, instructors, staff, and teaching assistants (collectively referred to as the graders) are seated in a circle, with each grader assigned to grade a single station of the exam. Each grader is given an answer key and a stack of student tests. If a grader is unsure of an answer, they record that questionable answer on a separate piece of paper. Once each grader is finished with the unquestionably correct and incorrect answers, they each take a turn reading aloud their list of questionable answers. The group holds a discussion on each questionable answer, where everyone is encouraged to state their case for that answer being correct or incorrect. After the discussion, the group votes and the majority decision is enacted. Any answers designated as correct by this democratic process are added to the answer key while incorrect answers are added to an "incorrect answer key", alongside the rationale as to why the answer was voted to be incorrect. As the process unfolds, fewer and fewer questionable answers arise as both keys grow. This process is beneficial because it allows for a diverse range of expert input, in tandem with a democratic process, to arrive at what is hopefully the best possible answer key. Drawbacks of this process include the demand for many graders as well as an increased time to completion, as deliberations can sometimes become heated and drawn out.

**Decision Trees**

Decision Trees (DTs) have been found to be effective at parsing grammatical structures in NLP (Leacock & Chodorow, 2003). For this research study, DTs were used to produce a set of rules that describe a relationship between the words in an OPSE answer and correctness. The following description of DTs is summarized from Quinlan unless otherwise noted (Quinlan, 1986).

The goal of decision tree learning is to minimize the entropy in the data set by maximizing the information gain with each choice made. All nodes in a tree have an associated classification label and probability that the label is correct (the method for assigning these is described below). A node is referred to as a "leaf node" or a "terminal node" if no other nodes stem from that node. All non-leaf (Quinlan, 1986) nodes in the tree represent a decision or Boolean (true/false) rule that splits the data in two, based on a feature in the data set. A schematic of an example DT is shown in Figure 1. The rule that gains the most amount of information is chosen for each node. To determine which rule gains the most information, the data set is split for each unique value of every feature within the data set. The information gain is then calculated for every possible rule. A rule that is being considered is referred to as the "candidate rule". If C and I are the counts for the number of correct and incorrect results in the data, respectively, then entropy (E) is computed as:

$$E(C, I) = -\frac{C}{C+I} \log_2 \frac{C}{C+I} - \frac{I}{C+I} \log_2 \frac{I}{C+I} \qquad \text{(Harris et al., 2020)(Equation 1)}$$

Grading Anatomical OSPE Using Decision Trees

If the data set consists of *x* features and each feature has *y* unique values, then the entropy for each feature's unique value is denoted as $E_{x,y}$. The entropy value is computed by first splitting the data into two subsets: one for which the candidate rule is true ($S_T$), and another subset for which it would be false ($S_F$). For "$S_T$", Equation 1 is used to compute $E_{x,y,T}$, with C and I being the counts for correct and incorrect results in $S_T$. The same is done for $S_F$ to compute $E_{x,y,F}$. $E_{x,y}$ is the weighted (by the relative sizes of $S_T$ and $S_F$) sum of $E_{x,y,T}$ and $E_{x,y,F}$. Information gain (IG) is computed as:

$$IG_{x,y} = E_{curr} - E_{x,y} \qquad \text{(Equation 2)}$$

Where $E_{curr}$ is the current entropy for the current data set and $E_{x,y}$ is the entropy for each *x* and *y*. The best choice is the rule for which the greatest value of information gain is calculated (See Equation 3).

$$IG_{Best} = \operatorname*{argmax}_{1 \le i \le x,\, 1 \le j \le y} IG_{i,j} \qquad \text{(Equation 3)}$$

Once a rule has been selected for the root (first/top) node of a tree, the left subtree is built using $S_T$, and the right subtree is built using $S_F$. Since the nodes that stem from the root node are themselves the roots of their respective subtrees, the process simply repeats recursively until 1) no split of the data can be performed (this occurs when all samples have uniform values) or 2) all samples have the same classification. The data set used by a node to determine the rule also determines the associated classification label and probability for that node. The classification label is the classification with the largest count within the data set used for that node. This count decreases as the algorithm goes deeper into the tree. The probability of the input being correct is the proportion of the classification label to the size of the data set used for the node.

Classifying a new sample using a trained DT is achieved by applying a rule to the sample (starting with the top root rule), and then traversing left or right based on whether the rule computes to true or false. The classification terminates when a leaf node is reached, or when the rule of the node does not apply. The latter can occur when the new sample contains features that were not present in the training set. The resulting classification is the label at which the node terminates, and the certainty is the associated probability at the terminal node.

MATERIALS AND METHODS

**Online OSPE**

For the winter online OSPE, two-dimensional images from the Stereoscopic Atlas of Human Anatomy (Bassett, 1962) were used in place of cadaveric material. The images were edited to include a digital pin that denoted a specific anatomical structure that formed the basis of the station's line of questioning. The 18 images and 54 questions (3 questions per image) were uploaded to the learning management system (LMS) Desire to Learn (D2L), which at McMaster University is called Avenue to Learn (Avenue). The quiz feature in Avenue was unable to recreate the individually timed stations present in an in-person OSPE. Therefore, students were given a total of 36 minutes (2 minutes per station) to complete the entire examination. Answers were submitted by typing into a text box on the webpage. An online practice OSPE, comprising 5 images and 15 questions, was given to students to become accustomed to the online format. To minimize academic integrity concerns, the order of the stations was randomized for each student. In addition to the station randomization, each station was restricted to a single page within the Avenue website and students could not navigate to previous stations. This presented an issue, as the station randomization caused the answer key to be out of sync with each student's test. Therefore, a means of sorting questions was required. To be able to sort the questions, six-digit identifiers were appended to each question text. To curtail collaboration during the exam, the identifiers were obscured from student view by changing their font colour to the webpage background colour. An example of an online OSPE station is shown in Figure 2. For this study, the answers submitted by students and the associated grades from 368 OSPE submissions were used.

The grading, discussion, and democratic process for the winter semester online OSPE included two anatomy faculty members and two staff members with extensive experience in anatomy education, collectively referred to as the expert graders. As the meeting could not be done in-person, the group was kept small as it was easier to manage online. However, sufficient expertise remained present throughout the group to ensure a high probability that the grading was accurate. When the voting on the correctness of a student's answer resulted in a tie, the answer was deemed correct and subsequently added to the answer key.

**Grading OSPEs with Decision Trees**

The data set that was used by the DT constituted all unique student-submitted answers. No misspellings were corrected in either the correct or incorrect answers, which is a notable difference from the TA process where the TA can apply some judgement on the intent of the student. Some consideration was given to using a spell checker on the data set. However, for this research, it was rejected because it introduces spellchecker accuracy as an additional variable. Additionally, from a practical standpoint, since there will always be some misspelled words that cannot be reliably fixed by spell checking, it was deemed best to investigate how misspelled words affect accuracy. The use of spell checking is a possible avenue for enhancing the algorithm in the future.

Prior to algorithm selection it was observed in the data set that, for most questions, the correct answers had little variety. This makes sense, as the questions are highly technical and the answers very specific. If a student knows the answer, there are very limited ways to express it properly. Since answers could not receive partial credit, if an answer was not correct, then it must be incorrect. While axiomatic and obvious, this is important for producing a good model of the answers, as it means the algorithm needs to be able to only identify correct answers. All other answers are incorrect, and no mechanism is required to model recognizing an incorrect answer *per se*. This suggested using an algorithm that uses a positive logic model, i.e., an answer is correct if and only if, it has the features of a correct answer (including "correct" words and excluding "incorrect" words); otherwise, it is incorrect. As a real-world example, this is like the logic that might be used for a fire safety system which activates if and only if there is positive evidence of a fire; otherwise, no fire is assumed.

Based on these observations, it was reasoned that a good algorithm for creating such a model was a decision tree (DT). After removing common words such as "the", "that", "a", "an", etc. there were generally only a few unique words remaining in the set of correct responses. Hence, while the possible lexicon of words that can appear in an answer is technically large, i.e., all words in a language, the resulting tree is quite small as the recursive call terminates when the remaining classifications are uniform. In other words, when an answer only contains words that classify to being incorrect, the tree stops growing. Additionally, the use of a DT allows for different possible combinations of correct words to appear together in the model, even if a specific combination has never appeared before as an answer. However, DT models are not exhaustive and do not define a complete set of all possible correct answers. Finally, a DT allows for an explanation of its reasoning and a certainty of correctness.

An explanation, including the importance of each rule in coming to a decision and the critical decision point, can be provided by parsing the rules of the tree for a given answer. Importance is the probability at each node when traversing the tree, and the critical decision point is the node with the greatest importance of all those traversed. Three examples are described as follows, with the corresponding anatomical structure shown in Figure 2 and the DT shown in Figure 3.

1. Student answer: Papillary Muscles (Marked correct by expert graders)

    First node "muscles" returns TRUE,

    Second node "papillary" returns TRUE.

    DT returns "answer is correct (97% significance)"

2. Student answer: Atrial Papillary Muscles (Marked incorrect by expert graders)

    First node "muscles" returns TRUE,

    Second node "papillary" returns TRUE,

    Third node "Atrial" returns TRUE.

    DT returns "answer is incorrect (100% significance)"

3. Student answer: Subvalvular Apparatus (Marked correct by expert graders)

    First node "muscles" returns FALSE,

    Second node "subvalvular" returns TRUE,

    Third node "apparatus" returns TRUE.

    DT returns "answer is correct (100% significance)"

Contrasting example 1 and example 2: the DT is only 97% confident in the correctness of "papillary muscles" because additional pieces of information, such as the word "atrial" shown in example 2, can drastically affect the correctness of an answer. In example 3, the answer fails to include the commonly correct word "muscles". However, it does include both words "subvalvular" and "apparatus" which allows it to follow an alternate path to correctness.

**Evaluation of the Decision Trees**

For this research, a separate DT was built for each question from the examination. For each question, a data set was constructed of all non-blank answers given by the students, and each answer was given a label of correct or incorrect exactly as per the grading of the answer by the grading team. The features of the data set for a question were the unique words across all answers, with the common English words: "a", "an", "and", "are", "as", "at", "be", "but", "by", "did", "for", "had", "has", "have", "I", "in", "is", "it", "of", "on", "or", "so", "than", "that", "the", "then", "they", "this", "to", "was", "with" removed.

For evaluating the algorithm, a 10-fold cross validation method was used, and as DTs are deterministically trained in a single step, no validation set was required. In other words, 90% of the data (X number of tests) were used for training the DTs and 10% (x number of tests) used for testing the DTs, with no answer appearing more than once in the testing set. For each fold, the DT was built using the training data and then each sample in the test data was classified. A count was kept of the number of answers that were assigned the proper classification label. The accuracy for each question was computed as the number of proper classifications over the total number answers.

Statistics

All statistical tests were performed using Matlab, R2020b, version 3.0 (Mathworks, Natick, Mass.).

RESULTS AND DISCUSSION

**DTs Grade Anatomy OSPEs with a High Degree of Accuracy**

The following section provides the results for the evaluation of using a DT to grade anatomy OSPEs. First, the accuracy results are given in Table 1 and discussed. No comparison is provided to existing approaches as they generally are intended for a broader category of short answers questions and for K-12 education (Burrows et al., 2015). Afterwards, the observations made while evaluating the algorithm are discussed, focusing on exploring the incorrect classifications which required looking at the number of unique words in the answers (shown in Table 2). This includes an examination of the reasons why the algorithm functions, with a comparison to other existing similar approaches. Additionally, Table 3 shows the correlation values for tests between the accuracy results and different features of the data.

Table 1 shows the average grade for each question followed by the accuracy of the DT, and the difference between them. In terms of overall accuracy, the DT averages 94.49% over all 54 questions. Most of the time the accuracy is quite high: only 7 of the 54 questions have an accuracy less than 90%, and only 2 of those 7 have an accuracy less than 80%. This suggests that the approach has some potential for grading OSPE questions with good accuracy. However, good accuracy is meaningless if it is found by guessing. A Pearson correlation value of 0.22 (shown in Table 3) indicates the values are not well-correlated to the grades; however, the p-value is 0.09. While $p > 0.05$, it seems unlikely that the DT is simply guessing as it was found to use good reasoning for grading an answer correct. Another test to ensure that the DTs are not guessing is to consider the null hypothesis described in the next paragraph. Additionally, by examining the explanations provided by the DT, they are generally sensible and well-justified. The DT finds that an answer is correct if it contains words that have been found in other correct answers, and it is able to express that logic.

One null hypothesis in this case is to classify every answer as correct, which in effect, is equal to the average grade. For example, if 75% of students answered a question correctly, and the null hypothesis model classified every student's answer to that question as correct, then the null hypothesis model would be 75% accurate. Similarly, if the null hypothesis model classifies every student's answer as incorrect, then when the grades are low the null hypothesis model would be more accurate. Hence, any significant effect from using the DT is best seen by examining questions where the students' grades are close to 50% as this should be the hardest for the DT to predict by only guessing without reasoning. In Table 1, the questions with an average grade between 40 and 60% are italicized, and from these rows it can be seen that the DTs have an average accuracy of 93.9% accuracy, which is 0.59% less accurate than the overall average.

**Common Sources of Errors with DT Grading**

While the DT had high accuracy, it does exhibit three common errors that could be areas for future study to further improve accuracy. As mentioned previously, the algorithm used in this research is based on the observation that the questions in OSPEs elicit correct answers that consist of technical language from students and a limited lexicon. For these types of questions, there are only so many possible ways to be correct. Hence, answers that contain these technical words are almost always graded correctly. In examining the cases where the algorithm marked improperly, three cases emerged as the most common source of error:

1. When a correct answer lacks specificity and shares a few words in common with incorrect answers. For example, the first follow-up question after identifying the papillary muscle in the OSPE station presented in Figure 2 (labeled as question 53) asks the students to identify the function of the papillary muscle. The student answer "contract av valves closed" is graded as correct by the DT model. However, this student answer is incorrect because answers like "contract av valves" misidentify the function of the papillary muscles. Pressure in the ventricles cause the atrioventricular valves to close, the papillary muscle prevents the atrioventricular valves from prolapsing. The DT likely classified the student answer "contract av valves closed" as correct due to other answers such as "keep av valves closed". The student answer "keep av valves closed" describes the concept of preventing prolapse and was, therefore, classified as correct by the expert graders. However, the student answer "keep av valves closed" also includes the word "closed", potentially resulting in the DT strengthening its connection between the inclusion of the word "closed" and the student answer being classified as correct. The nuance of "keeping valves closed" and "preventing prolapse" versus "closing an av valve" is missed by the DT but clear to the experts. The existence of words in both a correct and incorrect answer is confusing to the DT. For example, the DT was somewhat uncertain with the above classification, indicating only 30% certainty of being incorrect. In practice, situations like this would likely be audited by a human expert and would likely be found by the student during exam review. This issue could be further addressed by including additional natural language processing techniques to the algorithm, to help the DT understand that "contract" does not mean "to close" in this case.

2.  When an incorrect answer lacks specificity and shares many words in common with correct answers. For example, the answer "maintain tension on chordae tendinae" is graded as incorrect by the DT model since it does not include mention of the "atrioventricular valve", which the expert graders considered critical for the answer to be correct. As there were correct answers that included the words "chordae", "tendinae", "tension", and "atrioventricular valve" (or "av valve") during classification, the algorithm runs out of words to traverse the tree. Hence, it is at a node with a probability of being correct greater than 50%. For this example, the DT graded it as correct with only 55% certainty, which again would likely be audited and corrected as above.

3.  A unique correct answer. In some cases, a correct answer was so unique that it contained no similar words to other correct answers. These answers violate the assumption made by the initial observation that most correct answers use similar words, and hence could be graded incorrectly. These are fundamentally harder to deal with except by simply making a more complete answer key. In this case a low certainty in the answer from the DT would not be calculated and this could only be corrected if the error was detected during exam review.

**Higher DT Grading Accuracy Correlates with Lower Number of Unique Words Among Answers**

In examining the questions with accuracy less than 90%, it was observed that such questions tended to have a higher number of unique words (Table 2). A correlation value was computed for accuracy versus the number of unique words across all, only correct and only incorrect answers (Table 3). It was found that accuracy was negatively correlated to the number of unique words with Pearson correlation values of -0.71 (p= 0.07), -0.76 (p <0.01), and -0.63 (p, 0.01). The negative correlation between unique word number in the OSPE answer and high accuracy makes intuitive sense as the number of unique words will cause a larger decision tree, which leads to more unique paths that can cause confusion during classification by traversing a path that bears similarity to the answer but is not actually classified correctly. Therefore, it is best to craft questions that lend themselves to be easily answered in a succinct, technical manner. It appears that long; complex answers lead to less accuracy when using a DT in the fashion presented in this paper and that the DR is not a general solution to short answer questions. In other words, a DT-based algorithm works best when the questions have a limited number of ways to express a correct solution. The initial assumption is that this number should be low for a DT to function properly, so this reinforces that the initial assumption was correct. It suggests that automated examination design could be improved by crafting questions with this property in mind. However, it is unclear whether such questions are effective for reinforcing student learning.

Relative to other existing approaches, the DT performs well within this limited context. For example, C-rater, a tool for the automated grading of short narrative questions, has an accuracy for different concepts ranging from 55% to 94% (Sukkarieh and Blackmore, 2009). Additionally, Shermis et al. performed a competition comparing human and different machine learning approaches across different data sets (Shermis et al., 2015). They found that machine learning algorithms typically had a non-random agreement between the algorithm and expert raters (Cohen's kappa) in the 70-80% range. While the algorithm presented herein technically has higher accuracy, a direct comparison is not fair to these existing approaches that use far more sophisticated mechanisms to achieve more generalizable algorithms. A survey (Burrows et al., 2015) of existing approaches for automated short answer grading shows the history and complexity of this research area. However, the existing approaches do make similar observations to what was found in this research study. Leacock and Chodorow found that pronouns, negatives, misspellings, lack of specificity and aberrant answers were a particular issue (Leacock and Chodorow, 2003). For OSPE-style anatomical questions, pronouns and negatives seldom appear and therefore do not tend to be an issue but a lack of specificity with either correct or incorrect answers, as well as a high number of unique answers, could lead to incorrect classifications. Similarly, Shermis (Shermis, 2015) found that automated short answer grading using machine learning can be improved by limiting the lexicon to a finite pool of correct answers, which is the underlying hypothetical of this research. The results of this paper reinforces their findings, and suggest the approach herein was justified.

Study Limitations

- The results are for only one anatomy exam for one undergraduate group of students. Numerous exams over a number of years with a more varied population would doubtless improve the algorithm.
- Adding spellchecking or other sources of information for the AI (such as a medical dictionary) may improve accuracy as the AI currently only uses student answers as a source of information.
- The anatomical questions selected may not be representative of all the types of anatomy thus DT-based AI may work better or worse for other types of anatomy.

CONCLUSIONS

This paper presented an approach for grading Objective Structured Practical Examinations, colloquially referred to as "bell ringer" exams. These exams consist of a series of stations, each with a corresponding anatomical specimen or model. A student must write down a short answer, usually identifying a pinned anatomical structure or its anatomical function. It was hypothesized that the nature of such questions should result in correct answers that have a limited lexicon of unique words, as there are only so many ways to be correct. A preliminary cursory examination of the answer data seemed to support this notion. As such, a DT algorithm was selected to model the answers to the questions. DTs create models that consist of a series of Boolean rules, which are effective when the classification can be sharply divided. In this case, the correct answers were highly identifiable. Axiomatically, if the answer did not contain identifying features of the correct answers, then it is incorrect. Using this approach, the DT was able to grade the answers with 94.49% average accuracy across 54 questions, using the grading of expert graders as the correct classification.

While accurate, DT-based AI of marking OSPEs has some limitations. Most importantly, the OSPE answers need to have a small, technical lexicon of words. It can be seen in the results that as the number of unique words increase, accuracy decreases with a correlation value of approximately -0.7. This can be overcome by carefully designing the questions, but it would be better to improve the algorithm to be less sensitive to the number of unique words.

The training set, which consists of the answers from a cohort of 368 students, has multiple uses outside the scope of this experiment. The questions remain useable for subsequent exams and as intended with this research, in an intelligent tutoring system for future cohorts. Ideally, it would be helpful to be able to grade answers with only an answer key. This was not investigated in this research because the answer key was not designed with machine learning in mind. However, future investigations with a machine learning suitable answer key will be done. Overall, though, as a first step towards grading anatomical OSPEs, the approach herein shows considerable potential.

The main intended impact of this experiment is to work towards developing an intelligent tutoring system that can aid anatomy students in learning the material. To implement an intelligent tutoring system requires automated grading, since it is impractical to have a human grader available on demand. Automated grading also allows for exams to be graded without expert involvement, with the exception of auditing uncertain results from the AI tool. Other the other hand, from a pedagogical standpoint, it is helpful for the teaching assistants to grade the examinations to develop their skills. Therefore, it is envisioned that the AI tool will be used for confirmation as opposed to replacing teaching assistants.

In addition to the considerable effort required to generate a large cohort of answers, the main issue is related to how the tree is constructed. For this research, a standard "greedy" approach was used. In other words, the word with the most information gain was selected as the rule at each stage. However, this is not the only technique that can be used in constructing a DT. A "thrifty" approach considers options more holistically, sometimes selecting an option that seems less optimal, but overall leads to a better outcome. It is suspected that the "greedy" approach is responsible for creating trees that are susceptible to answers that lack specificity, as discussed earlier. Alternative tree construction techniques will be explored to address this weakness.

ACKNOWLEDGEMENTS

The authors wish to thank Courtney Pitt, the senior instructional assistant, for her contributions in setting up and troubleshooting the final OSPE on McMaster University's learning management system. The authors would also like to thank her for organizing the OSPE "marking bee" as well as her contributions to marking the OSPE.

The authors also wish to thank Dr. Alexander K. Ball, Professor Emeritus in the Department of Pathology and Molecular Medicine one of the course coordinators, for his contributions in editing the OSPE prior to administration, as well as his contribution to marking the OSPE exam.

NOTES ON CONTRIBUTORS

JASON BERNARD, Ph.D., is a postdoctoral fellow working at McMaster University in the Department of Surgery. He is on the program committee for the European Conference on Technology Enhanced Learning, and a review editor for Frontiers in Artificial Intelligence. He is a reviewer for several journals such as Computers and Education, Natural Computing, and Applied Soft Computing. His research interests are in using artificial intelligence for educational technology and model inference of natural processes, most recently in a medical context.

RANIL SONNADARA, Ph.D., holds academic appointments at McMaster University, the University of Toronto and the Vector Institute for Artificial Intelligence. He also serves as the Director of the Office of Education Science in the Department of Surgery at McMaster. His research interests focus on human performance in high stakes environments, competency-based models of education, and the role of technology in learning.

ANTHONY N. SARACO, M.Sc., B.Eng., Biosci., is a research assistant for the Education Program in Anatomy, Faculty of Health Sciences at McMaster University, Hamilton, Ontario Canada. He teaches gross anatomy and physiology to various undergraduate disciplines and is passionate about bridging the gap between engineering and medical sciences.

JOSH P. MITCHELL, B.Eng., is the developer at the Education Program in Anatomy in the Faculty of Health Sciences at McMaster University, Hamilton, Ontario Canada. He creates physical and digital solutions to facilitate research into the efficacy of various new education technologies, with a current focus on virtual reality and augmented reality.

ALEX B. BAK, B.H.Sc., is an M.D. candidate in the Faculty of Medicine and M.Eng. candidate in the Institute of Biomedical Engineering, both at the University of Toronto. He is also a researcher at the Education Program in Anatomy, McMaster University Faculty of Health Sciences. His research interests include innovating and testing technologically-enhanced tools and applying it to improve patient care and education. He is motivated to provide high-value care and education that is accessible and equitable.

ILANA BAYER, Ph.D., is an assistant professor of pathology and molecular medicine, and the Director of the Faculty of Health Sciences Learning Technologies Lab at McMaster University, Hamilton, Ontario, Canada. She teaches anatomy and physiology at the undergraduate level and teaches on the use of technology in health science education and healthcare to various programs ranging from undergraduate level to faculty development. Her research interests focus on the use of technology in higher education.

BRUCE C. WAINMAN, Ph.D., F.A.A.A is a professor of pathology and molecular medicine, and the Director of the Education Program in Anatomy and Surgical Skills Laboratory at McMaster University, Hamilton, Ontario, Canada. He teaches anatomy, physiology, and pharmacology to various programs at the undergraduate and graduate levels. His academic interests are in the role of extended reality in anatomy education, interprofessional education and the ethics of body donation.

Grading Anatomical OSPE Using Decision Trees

FIGURE LEGENDS

**Figure 1: The Anatomy of a Decision Learning Tree:** A Decision Learning Tree (DLT) is a series of nodes containing unique words that are connected by Boolean (True/False) decisions. The nodes are described as either a "root node" (a node that has nodes stemming from it) or a "leaf/terminal node" (a node that does not have nodes stemming from it). All nodes that result from the Boolean decision returning "True" are included in the left subtree (surrounded by blue) while all nodes that result from the Boolean decision returning "false" are included in the right subtree (surrounded by pink). The asterisks (*) are a "wild card", representing any word that is not one of the unique words. In this case a correct OSPE

answer was atrial papillary muscle(s) or subvavular apparatus and all other answers would have been found to be false by the DLT.

**Figure 2: Objective Structured Practical Examination (OSPE) Question Example:** Example of a station and question set used in the winter semester online OSPE. The six-digit identifier was obscured from the students' view. Students were aware that use of the term "question (A)" referred to the first question of the station.

**Figure 3: Decision Tree Example:** A decision tree structure where the correct answers only contained the words "muscles", "papillary", "subvalvular", and "apparatus". The percentage indicates the percentage of answers that were incorrect that passed through that node. For classifying an answer, a left arrow indicates that the answer contained the word within the node (Boolean returning true). A right arrow indicates that the answer did not contain the word within the node (Boolean returning false). An asterisk indicates all possible words.

Grading Anatomical OSPE Using Decision Trees

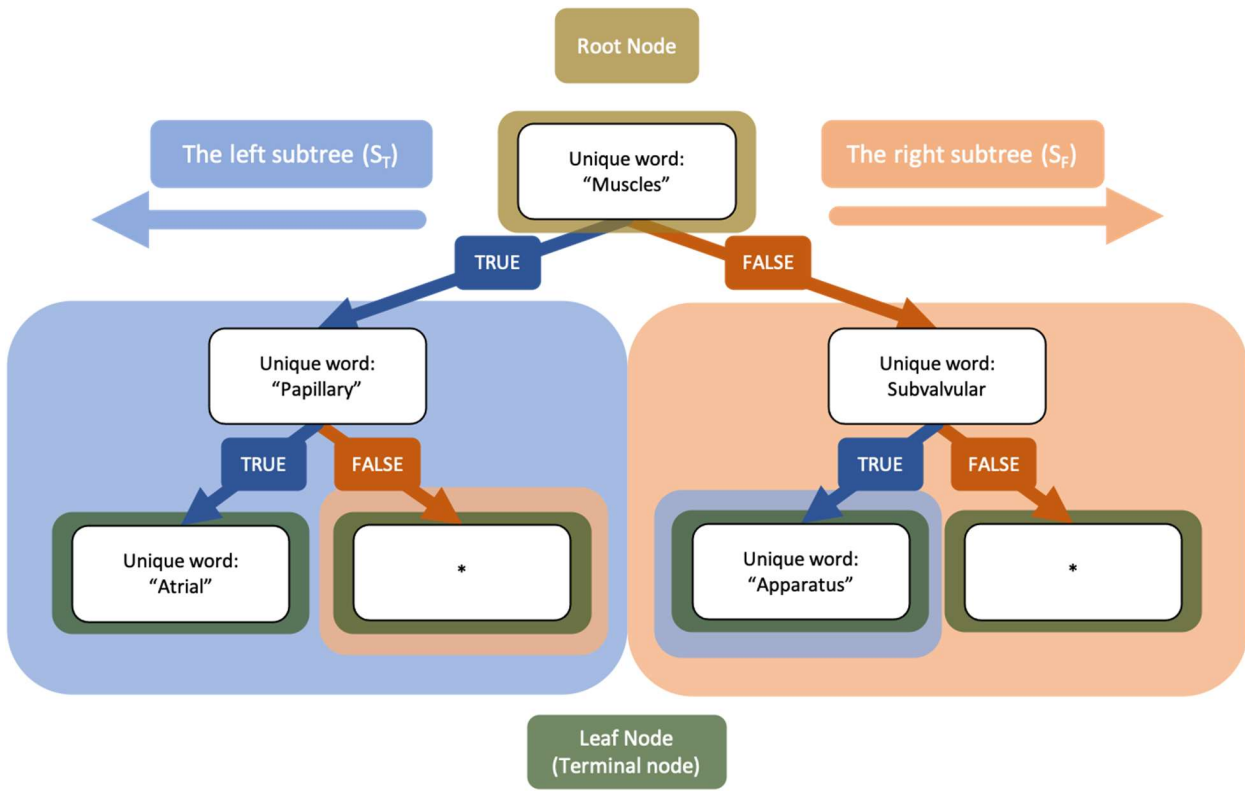

Figure 1: The Anatomy of a Decision Learning Tree: A Decision Learning Tree (DLT) is a series of nodes containing unique words that are connected by Boolean (True/False) decisions. The nodes are described as either a "root node" (a node that has nodes stemming from it) or a "leaf/terminal node" (a node that does not have nodes stemming from it). All nodes that result from the Boolean decision returning "True" are included in the left subtree (surrounded by blue) while all nodes that result from the Boolean decision returning "false" are included in the right subtree (surrounded by pink). The asterisks (*) are a "wild card", representing any word that is not one of the unique words. In this case a correct OSPE answer was atrial papillary muscle(s) or subvavular apparatus and all other answers would have been found to be false by the DLT.

Grading Anatomical OSPE Using Decision Trees

**Question 52** (1 point)

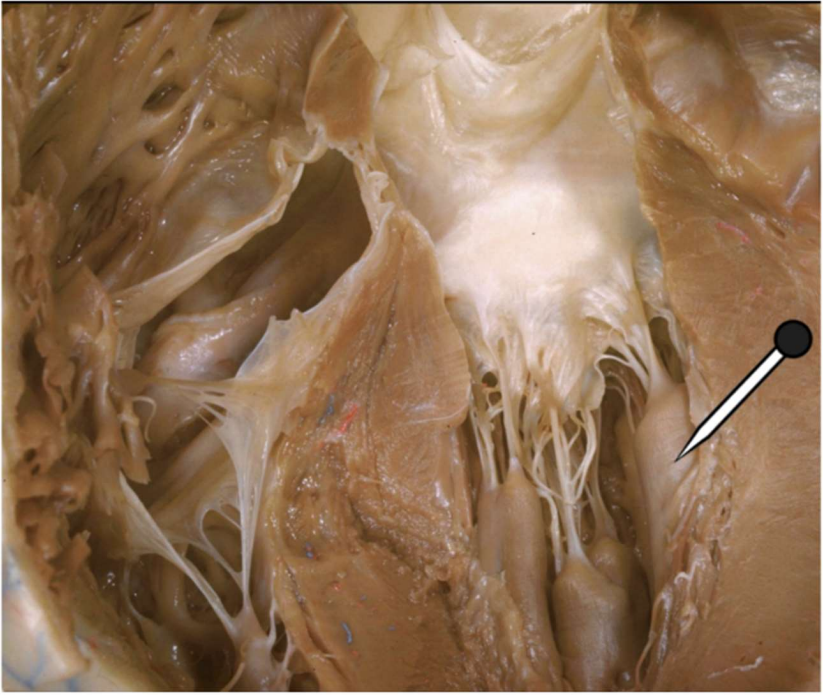

Identify the structure indicated by the black pin.

**Question 53** (1 point)

What is the function of the structure indicated by the black pin?

**Question 54** (1 point)

What is responsible for signaling depolarization of the structure identified in question (A).

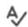

Figure 2: Objective Structured Practical Examination (OSPE) Question Example: Example of a station and question set used in the winter semester online OSPE. The six-digit identifier was obscured from the students' view. Students were aware that use of the term "question (A)" referred to the first question of the station.

Grading Anatomical OSPE Using Decision Trees

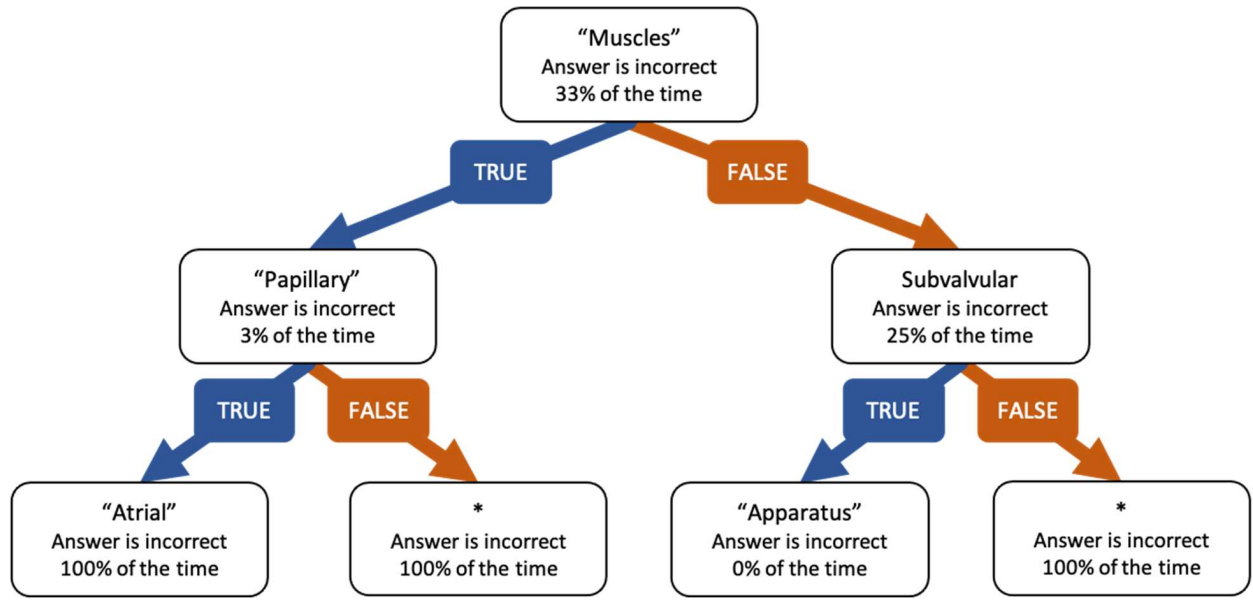

Figure 3: Decision Tree Example: A decision tree structure where the correct answers only contained the words "muscles", "papillary", "subvalvular", and "apparatus". The percentage indicates the percentage of answers that were incorrect that passed through that node. For classifying an answer, a left arrow indicates that the answer contained the word within the node (Boolean returning true). A right arrow indicates that the answer did not contain the word within the node (Boolean returning false). An asterisk indicates all possible words.

Table 1: Accuracy of the Decision Learning Tree. For each question in the test set, the average grade achieved by the students on the examination and accuracy of the DLT in grading the question is shown. The average grade. The seven questions with accuracy less than 90% are bolded. Questions with an average grade between 40 and 60% are italicized.

| Question | Average Grade | DLT Accuracy | Question | Average Grade | DLT Accuracy |
|---|---|---|---|---|---|
| Q1 | 0.951 | 0.989 | Q28 | 0.628 | 0.971 |
| **Q2** | **0.753** | **0.814** | Q29 | 0.747 | 0.983 |
| Q3 | 0.826 | 0.980 | Q30 | 0.323 | 0.947 |
| Q4 | 0.617 | 0.952 | *Q31* | *0.462* | *0.974* |
| **Q5** | **0.128** | **0.873** | Q32 | 0.299 | 0.953 |
| Q6 | 0.663 | 0.979 | *Q33* | *0.495* | *0.970* |
| Q7 | 0.769 | 0.960 | Q34 | 0.641 | 0.929 |
| Q8 | 0.734 | 0.943 | Q35 | 0.378 | 0.956 |
| *Q9* | *0.410* | *0.942* | Q36 | 0.524 | 0.950 |
| *Q10* | *0.486* | *1.000* | Q37 | 0.840 | 0.981 |
| Q11 | 0.226 | 0.938 | *Q38* | *0.408* | *0.947* |
| Q12 | 0.147 | 0.906 | **Q39** | **0.614** | **0.850** |
| *Q13* | *0.443* | *0.969* | Q40 | 0.668 | 0.982 |
| Q14 | 0.188 | 0.962 | *Q41* | *0.489* | *0.985* |
| Q15 | 0.171 | 0.977 | *Q42* | *0.514* | *0.944* |
| Q16 | 0.766 | 0.957 | Q43 | 0.375 | 0.988 |
| Q17 | 0.356 | 0.921 | Q44 | 0.342 | 0.953 |
| Q18 | 0.810 | 0.994 | Q45 | 0.296 | 0.931 |
| Q19 | 0.745 | 0.966 | *Q46* | *0.457* | *0.969* |
| ***Q20*** | ***0.478*** | ***0.891*** | Q47 | 0.576 | 0.967 |
| ***Q21*** | ***0.451*** | ***0.734*** | Q48 | 0.709 | 0.981 |
| Q22 | 0.829 | 0.977 | Q49 | 0.342 | 0.976 |
| Q23 | 0.717 | 0.906 | Q50 | 0.247 | 0.952 |
| Q24 | 0.448 | 0.985 | **Q51** | **0.296** | **0.773** |
| Q25 | 0.584 | 0.977 | Q52 | 0.603 | 0.953 |
| Q26 | 0.774 | 0.997 | Q53 | 0.345 | 0.955 |
| **Q27** | **0.649** | **0.885** | Q54 | 0.296 | 0.906 |

*Table 1: Unique Word Count.* The number of unique words (after preprocessing) among all answers, only correct answers, and only incorrect answers. Questions with an accuracy less than 90% are bolded.

| Question | # of Unique Words | | | Question | # of Unique Words | | |
|---|---|---|---|---|---|---|---|
| | All | Correct | Incorrect | | All | Correct | Incorrect |
| Q1 | 36 | 24 | 16 | Q28 | 45 | 21 | 34 |
| **Q2** | **174** | **138** | **91** | Q29 | 33 | 14 | 21 |
| Q3 | 128 | 85 | 60 | Q30 | 72 | 47 | 42 |
| Q4 | 50 | 29 | 28 | Q31 | 78 | 20 | 67 |
| **Q5** | **285** | **86** | **263** | Q32 | 101 | 17 | 90 |
| Q6 | 40 | 19 | 25 | Q33 | 131 | 21 | 117 |

| Q7  | 57  | 30  | 28  | Q34 | 66  | 21  | 57  |
|-----|-----|-----|-----|-----|-----|-----|-----|
| Q8  | 82  | 47  | 45  | Q35 | 153 | 28  | 132 |
| Q9  | 190 | 98  | 121 | Q36 | 211 | 51  | 175 |
| Q10 | 32  | 5   | 32  | Q37 | 46  | 16  | 35  |
| Q11 | 170 | 46  | 145 | Q38 | 156 | 90  | 114 |
| Q12 | 156 | 52  | 117 | **Q39** | **272** | **190** | **155** |
| Q13 | 61  | 27  | 44  | Q40 | 50  | 14  | 38  |
| Q14 | 73  | 7   | 73  | Q41 | 89  | 15  | 78  |
| Q15 | 42  | 17  | 35  | Q42 | 217 | 64  | 178 |
| Q16 | 64  | 29  | 47  | Q43 | 105 | 17  | 90  |
| Q17 | 142 | 31  | 126 | Q44 | 113 | 29  | 88  |
| Q18 | 23  | 10  | 14  | Q45 | 309 | 71  | 264 |
| Q19 | 34  | 14  | 28  | Q46 | 113 | 33  | 104 |
| **Q20** | **179** | **75** | **134** | Q47 | 78  | 29  | 54  |
| **Q21** | **245** | **136** | **178** | Q48 | 67  | 18  | 55  |
| Q22 | 37  | 10  | 32  | Q49 | 97  | 21  | 84  |
| Q23 | 100 | 39  | 77  | Q50 | 87  | 15  | 80  |
| Q24 | 94  | 9   | 87  | **Q51** | **257** | **89** | **229** |
| Q25 | 50  | 9   | 44  | Q52 | 58  | 19  | 48  |
| Q26 | 42  | 11  | 32  | Q53 | 98  | 19  | 90  |
| **Q27** | **235** | **159** | **141** | Q54 | 120 | 29  | 95  |

**Table 2: Correlation Tests.** Results from correlation tests performed between accuracy and different features of the data

| Premise | Pearson Correlation |
|---|---|
| Null Hypothesis/Average Grade | $r = 0.22$, $p = 0.09$ |
| of Unique Words (All) | $r = -0.71$, $p = 0.07$ |
| # of Unique Words (Correct Only) | $r = -0.76$, $p < 0.01$ |
| # of Unique Words (Incorrect Only) | $r = -0.63$, $p < 0.01$ |

Grading Anatomical OSPE Using Decision Trees